\documentclass[sigconf]{acmart}
\AtBeginDocument{%
  }

\setcopyright{none}
\acmYear{}
\acmDOI{}
\acmConference[Thesis]{}{BSc in Computer Science}{July 2024}
\acmISBN{}

\usepackage{subfigure}

\begin{document}

\title{Hybrid Spiking Neural Network - Transformer Video Classification Model}

\author{Aaron Bateni}
\email{aaron.bateni@ut.ac.ir}
\affiliation{%
  \institution{University of Tehran}
  \city{Tehran}
  \state{Tehran}
  \country{Iran}
}

\author{Mohammad Ganjtabesh}
\authornote{The supervisor of the thesis}
\email{mgtabesh@ut.ac.ir}
\affiliation{%
  \institution{University of Tehran}
  \city{Tehran}
  \state{Tehran}
  \country{Iran}
}


\begin{abstract}
  In recent years, Spiking Neural Networks (SNNs) have gathered significant interest due to their temporal understanding capabili- ties. This work introduces, to the best of our knowledge, the first Cortical Column like hybrid architecture for the Time-Series Data Classification Task that leverages SNNs and is inspired by the brain structure, inspired from the previous hybrid models. We introduce several encoding methods to use with this model. Finally, we develop a procedure for training this network on the training dataset. As an effort to make using these models simpler, we make all the implementations available to the public at {\color{blue}\href{https://github.com/TheRNB/HyTSSN/tree/main}{github/thernb}}\footnote{https://github.com/TheRNB/HyTSSN}.
\end{abstract}



\settopmatter{printacmref=false}
\maketitle

\section{Introduction}
Conducting this research has been challenging, primarily due to
the nascent nature of these networks. With only a handful of pio-
neers in this domain, we often find ourselves navigating uncharted
territory but with highly promising results. These results highlight
the immense potential of this type of neural network to completely
replace classical neural networks and improve their performance.

The main objective of this research is to design and evaluate a
hybrid-model architecture inspired by the structure of the human
brain and apply it to classification tasks. The design of this archi-
tecture draws from prior hybrid-model architectures, but its core
element is inspired by and implements the cortical column struc-
ture of the brain. These columns, which are generally the primary
enablers of parallel processing in the brain, play a crucial role in its
accuracy and speed in responding to visual queries.

Through some experiments, we demonstrate the models promising
results. By analyzing these results, we point out both the high po-
tential and the limitations of the proposed approach and suggest
directions for future research to further optimize it for larger-scale
applications. But as our main goal, we provide the complete imple-
mentation of the model to be used in future work.

This work has been submitted to the Academic Council of the
Faculty of Science as a requirement for graduation at the under-
graduate level. It is hoped that these findings contribute to the
growing body of knowledge in this emerging field and inspire fur-
ther exploration of spiking neural networks at larger scales. The
code is also available on GitHub.

\section{Related Work}
With the introduction of the Transformer model in 2017 by the Google Research Team a revolution occurred in natural language processing.
\cite{fourth}
It did not take long for these models to be applied to visual problems for the first time in 2021 by researchers at Google.
\cite{fifth}
Initially, these models did not perform well, and convolutional neural networks remained dominant in image-centric problems. However, these models underwent significant changes that improved their performance. Notable among these changes were scaling the size of the models and modifying the core structures, such as the Swin Transformer architecture. 
\cite{sixth}
Alongside these optimizations, efforts were made to tackle other challenges in the realm of visual problems, leading to methods for converting video problems into image problems. 
\cite{seventh}
However, this still did not allow this architecture to outperform deep convolutional networks. 
\cite{eighth}
To address this, more complex models, such as the Video Swin Transformer
\cite{ninth} 
and the Mutliview Transformer
\cite{tenth}
were introduced.
These advancements allowed these models to match the accuracy of leading models in image and video recognition, which had until then been largely based on deep convolutional networks. However, a breakthrough in accuracy occurred when researchers considered an intriguing idea.

By this time, the working mechanisms and definitions of convolutional networks and transformers were well understood by the majority of the scientific community. To elaborate, convolutional models, due to the nature of the convolution function, create representations of closely related information in images and videos. In contrast, the attention mechanism in transformers is designed to capture representations of long-range or globally related information in images and videos. The research on this subject is strong and several models were introduced. Guo et al. introduced their model by feeding their input image into a couple of convolution layers, and then using it as the transformer's input. 
\cite{eleventh}
This, resulted in an input that already had the local information encoded in itself and all the transformer had to do was to find the correlation between the long range information. Tang et al.
\cite{second}
explored the idea of combining these two models to design a stronger hybrid model. Their model operated by running a small Swin Transformer and a small U-shaped convolutional neural network in parallel on a single input, combining their outputs to create a representation that included both global features and local features from the image. These hybrid models were the first to surpass the capabilities of deep convolutional networks in image-related problems, marking a new achievement.

While convolutional networks are very powerful in this domain, their fundamental limitation becomes apparent in video-related problems. Unlike images, which contain only visual information, videos contain two types of information: visual and temporal. For instance, in a video of an athlete running, both the spatial context in which the athlete runs and the act of running over time provide information. Since current networks do not inherently account for the temporal dimension, existing methods for video classification have relied on selecting multiple frames over time, devising methods to stitch these frames together into a larger composite frame, and then processing this composite as if it were a single image.
\cite{seventh, ninth, tenth, eleventh, first}

In recent years, a new type of deep neural network, called deep spiking neural networks, has been introduced and rapidly gained traction. The building blocks of these networks are neurons that, unlike current mathematically driven neurons such as those in the perceptron model, operate in the temporal dimension by mimicking the behavior of real brain neurons. For these networks, the timing of an input to a neuron becomes a critical factor.
\cite{twelfth}
 By replacing mathematically driven neurons with these temporally-aware neurons, classical networks can be made capable of recognizing and differentiating temporal dimensions in their input. Additionally, Kharadpisheh et al.
\cite{first}
demonstrated that these networks could also perform effectively in image-related problems.\\

\textbf{HTC-Net:} This paper
\cite{second}
addresses the challenge of improving accuracy and stability in disease diagnosis through medical imaging, a critical task for diagnosis and treatment decisions that rest on clinicians. Specifically, it focuses on the segmentation of skin lesions and intestinal polyps vital tasks for early detection and treatment in dermatology and gastroenterology. The datasets used are the publicly available ISIC 2017 dataset for skin lesion segmentation and the Kvasir-SEG dataset for intestinal polyp segmentation, both commonly used benchmarks in the medical imaging domain due to their importance in distinguishing pathological areas from healthy structures in imaging for disease prevention, such as cancer.

The HTC-Net model combines components from pre-trained deep convolutional neural networks and Swin transformers into a U-shaped encoder-decoder architecture. This hybrid approach leverages the ability of convolutional networks to extract fine-grained local features and the ability of transformers to process global features. The encoder employs pre-trained EfficientNet convolutional blocks and Swin transformer blocks to extract local and global features, while the decoder maps detected pathological regions back onto the image using convolutional neural network outputs and attention mechanisms of transformers. This innovative integration allows the model to effectively handle various scales and complexities in medical images, achieving more precise responses.

Results show that HTC-Net outperforms several advanced methods on the ISIC 2017 and Kvasir-SEG datasets. For skin lesion segmentation, HTC-Net achieved a Dice Similarity Coefficient of $90.07\%$ and an Intersection over Union of $84.02\%$, surpassing models like U-Net. Similarly, for intestinal polyp segmentation, the model achieved a Dice Similarity Coefficient of $91.18\%$ and an Intersection over Union of $85.94\%$. Comparison images in the paper demonstrate that HTC-Net produces more accurate and stable predictions, particularly in challenging cases with small or large lesions. This highlights the effectiveness of combining pre-trained convolutional and transformer models to achieve superior accuracy—a concept we aim to explore in this project.\\

\textbf{STDP-SNN:} This paper
\cite{first}
examines how to create a deep spiking neural network capable of high-accuracy object recognition. It focuses on using spike time-dependent plasticity (STDP) for unsupervised learning of object features, akin to how brain neurons adapt based on spike timing. The network’s ability to recognize objects despite changes in viewpoint, lighting, and occlusion is then evaluated using two datasets: the Caltech Face/Motorbike dataset and the ETH-80 dataset.

The proposed model consists of three convolutional layers, each followed by a pooling layer, progressively extracting more complex features from input images. The first layer detects simple edges, the second identifies distinctive object parts, and the third recognizes entire objects. An innovative aspect of the model is using STDP for unsupervised learning, allowing the timing of neuron spikes to influence synaptic strength. This biologically inspired mechanism enables the network to learn from temporal patterns in input data, reducing reliance on large labeled datasets. Additionally, the final layer employs global max pooling to enhance data invariance and ensure that the most prominent features are used for classification. A linear support vector machine is then employed for object category classification.

The network demonstrated promising results. On the Caltech Face / Motorbike dataset, it achieved a high accuracy of $99.1\%$, indicating strong generalization from training to testing. On the ETH-80 dataset, it achieved an accuracy of $82.8\%$, outperforming several other methods, including convolutional spiking neural networks and pretrained AlexNet. Notably, the network maintained robust performance even with limited training samples, achieving $95.1\%$ accuracy with only 40 images per category. Additionally, the model’s noise resilience was exemplary, maintaining reasonable accuracy at noise levels up to $20\%$. These results demonstrate that unsupervised STDP-based learning enables a deep spiking network to extract meaningful features from input data, making it a promising approach for efficient and effective object classification.

\section{Method}
\subsection{Preliminaries}
The primary question examined in this research is the introduction of the time dimension into problems solvable by classical neural networks. Traditionally, classical neural networks are composed of neurons performing mathematical computations where the time parameter plays no role. As soon as data is input into the network, computations are performed instantaneously, and the response is immediately available to the user. However, this structure is an incorrect simplification for data where the time dimension is present. For example, in continuous data like video, language sentences, or sound waves, the time of data entry into the network must be considered. Various tricks are employed to address this shortcoming in classical networks. However, the direct use of such data had not been previously proposed.

\subsubsection{Spiking Neural Networks}
Over the past century, the introduction of spiking neural networks revolutionized this field. Unlike classical networks, which are based on neurons with mathematical computations and immediate responses, spiking networks incorporate the time parameter into neuron activity. In this way, each neuron not only retains the input value but also the time of input. If two inputs do not occur simultaneously, they do not aid in activating the neuron. On the other hand, a neuron can activate multiple times over the course of receiving its inputs. Consequently, instead of being simply on or off, neurons exhibit an activity pattern referred to as a "spike."

\subsubsection{Cortical Columns of the Brain}
Cortical columns are fundamental units of classification in the brain's cortex, playing a critical role in information processing. The cortex, a gray outer layer of the brain, is responsible for various higher-level functions such as perception, cognition, and motor control. Within this precise and intricate structure, cortical columns are vertical clusters of neurons spanning six layers. Each column acts as a micro-organism capable of independently processing specific types of information, such as sensory inputs from the environment or internal cognitive processes.\\
The concept of cortical columns was first introduced by Vernon Benjamin Mountcastle in the 1950s based on his observations of the somatosensory cortex. He noticed that neurons within a small vertical volume of the cortex responded to the same specific type of sensory input, leading to his proposed theory of the organizing principle for cerebral function. Subsequent research has shown that this columnar organization is a recurring feature in different cortical regions, including the visual and auditory cortices. Current findings suggest that each column, containing approximately 80,000 to 100,000 neurons, processes a specific aspect of information. The aggregation of these outputs enables the brain to interpret complex data. The theory of cortical columns offers an explanation for the brain's parallel processing capability, which inspired our computational model aimed at emulating this biological parallel processing mechanism.
\cite{third}

\subsubsection{Spike Time-Dependent Plasticity}
Spike Time-Dependent Plasticity (STDP) is a biologically inspired learning algorithm that plays a critical role in neuroscience and neural network modeling. It represents a type of synaptic plasticity that adjusts synaptic connection strengths (analogous to connection weights in classical networks) over time in response to the activities of the pre- and post-synaptic neurons. What sets this method apart from classical approaches is its reliance on the precise timing of spikes.\\
The fundamental principle of the algorithm is based on the relative timing of spikes between the pre-synaptic neuron (spike from the sending neuron) and the post-synaptic neuron (spike from the receiving neuron). If a pre-synaptic spike occurs before a post-synaptic spike within a specific time window such that the pre-synaptic activity positively influences post-synaptic activity, the synapse strengthens—a process known as long-term potentiation. Conversely, if the post-synaptic spike precedes the pre-synaptic spike in a way that the post-synaptic activity is unrelated to the pre-synaptic spike, the synapse weakens, known as long-term depression. This time-dependent mechanism allows neurons to adjust their connections in a manner that enhances the network's ability to learn temporal sequences and patterns, such as continuous visual data, language sentences, or sound waves.

\subsection{Network Components}
By reviewing the model presented in \cite{second}, we can clearly see that a transformer has been used as a tool to improve the response of a deep convolutional network. In other words, if we intend to construct a hybrid model inspired by it, we must first build an independent neural network model, which is eventually combined with a transformer model, and their responses are fed to a shallow layer to obtain a final result. Since the structure of the transformer and the shallow layer has been developed by \cite{second}, we focus only on designing our neural network.

\subsubsection{Neuronal Groups}
\begin{figure}
	\begin{center}
		\includegraphics[height=8cm]{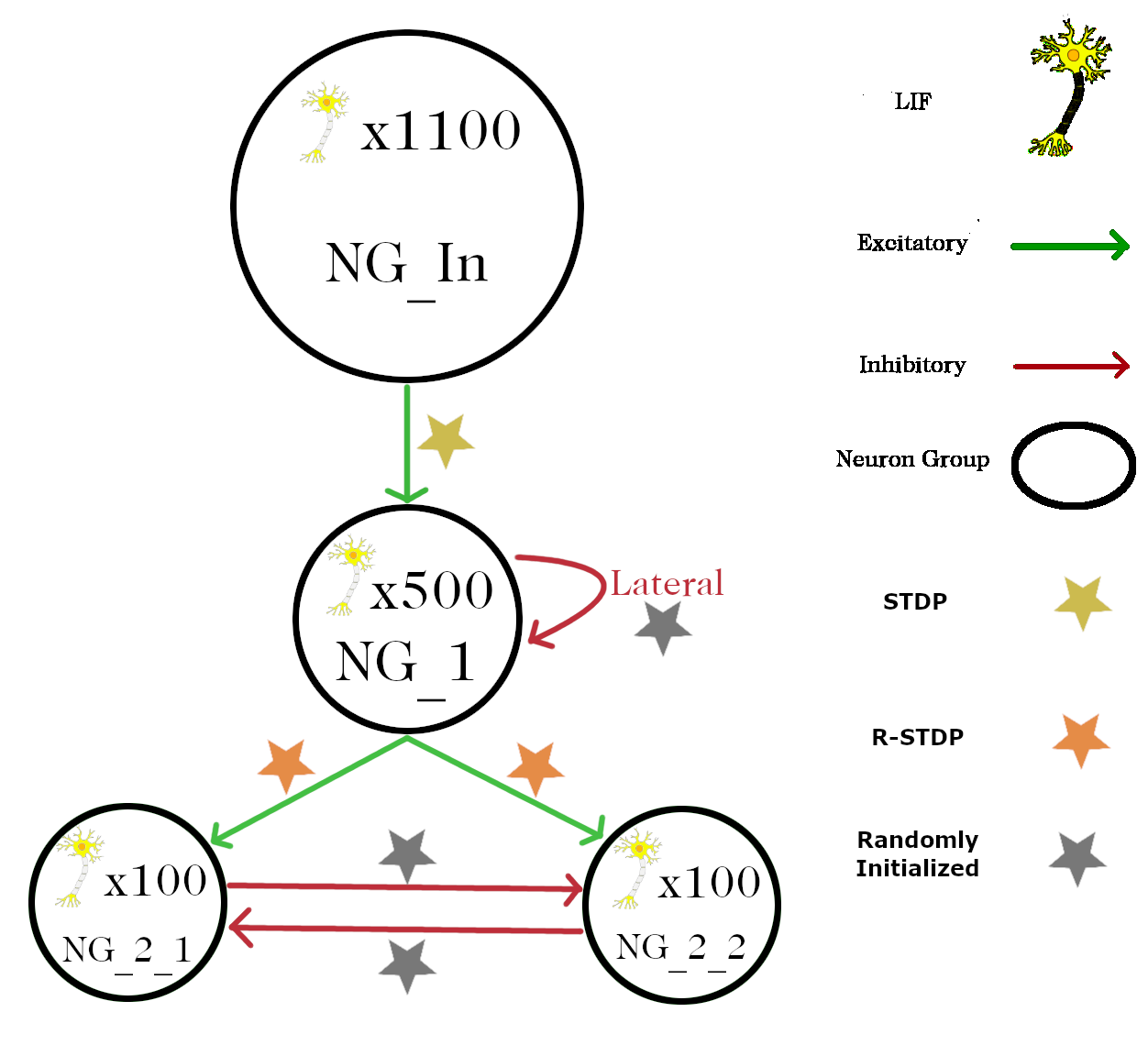}
	\end{center}
	\caption{
		View of the spiking network used for the mentioned classification tasks along with the learning mechanisms applied to each connection
	}
    \label{fig_architecture}
\end{figure}

To this end, we designed the network shown in 
\ref{fig_architecture}. 
In this network, the input is first converted into a spike chain using a random correspondence encoding set and enters the network through $NG_{In}$. Inspired by the hierarchical function of the brain, the network initially processes this raw information in a smaller layer to convert it into a compact representation. Then, inspired by cortical columns in the brain, the second stage consists of two final layers. If the input belongs to the first class, the first layer exhibits higher activity; if it belongs to the second class, the second layer becomes more active. This allows the network's judgment about the input to be extracted.

\subsubsection{Synaptic Connections}
The network includes several excitatory and inhibitory synaptic connections between the layers, described as follows:\\
In the input layer, there is only one excitatory connection to transfer the spike chain to the next layers, as the input sentences have already undergone initial processing to generate the spike chain, and further processing at this stage is unnecessary.\\
In the first layer, there are both excitatory and inhibitory connections. The excitatory connection is responsible for transferring messages to the second layer, while an internal inhibitory connection is also present. This connection ensures that only the primary features of the input are active, while less significant features are suppressed. This function can somewhat be compared to the convolution operation in deep convolutional networks for feature aggregation.\\
In the second layer, there are two neuronal groups, each representing a cortical column for one of the classes. This layer only includes two inhibitory connections. These ensure that if one group begins to activate upon recognizing an input, it informs the other group that it has started its activity, indicating that the input likely belongs to the class represented by the active layer. The active layer then suppresses the other, preventing its activation. This not only speeds up the network's response but also makes its decisions more decisive.

\subsubsection{Nature of Neurons and Response Detection}
Finally, the activity of neurons in the final layer of the network is calculated, and the more active layer is selected as the response. To compute the activity of neurons, we use the simple neuronal activity formula described below:
\begin{equation}
	A(t)=\lim_{\triangle t\rightarrow0}\frac{1}{\triangle t}\times\frac{n_{act}(t;t+\triangle t)}{N}=\frac{1}{N}\times\sum_{j=1}^{N}\sum_{f}\delta(t-t_{j}^{f}) 
\end{equation}

Where $A(t)$ denotes the activity, $N$ is the population size, $\Delta t$ is size of the time window, $n_{act}$ counts the number of spikes in a given time window, and $\delta$ is an indicator function that turns $1$ when there is a spike.

Additionally, for simplicity, all neuronal groups utilize the leaky integrate-and-fire neuron model with the following formula:
\begin{equation}
	\tau\times\frac{dU_{membrane}(t)}{dt}=-U_{membrane}(t)+R\times I_{input}(t)
\end{equation}
Where $U_{membrane}$ is the potential of the neuron, and $R$, $\tau$ are resistance and conductivity constant of the neuron. $I_{input}$ is the current that is inputted to the neuron at each time frame. The parameters for these neurons are defined as:
\begin{equation*}
    \tau = 10,~R = 5,~U_{reset} = -75,~U_{rest} = -67
\end{equation*}
\begin{equation*}
    U_{threshold} = -37,~I_{input} = 0.7
\end{equation*}

Where $U_{reset}$ is the minimum voltage, $U_{threshold}$ is the maximum voltage before spiking, and $U_{rest}$ is the resting potential.

\section{Training Procedure}
For training this model, we utilize two methods: spike-timing-dependent plasticity and reward-modulated spike-timing-dependent plasticity in unsupervised and supervised modes. Recall the network structure from the previous section.

\subsubsection{Spike-Timing-Dependent Plasticity}
In this structure, there is no way to validate the input structure in group $NG_1$ of the first layer, as there is no right or wrong for this structure! Hence, among the two mentioned methods, only spike-timing-dependent plasticity without reward can be used to learn this intermediate representation.

\subsubsection{Reward-Modulated Spike-Timing-Dependent Plasticity}
However, in the neuronal groups of the second layer, we can determine the correct response. Therefore, in these layers, we can use reward-modulated spike-timing-dependent plasticity for learning. To implement this, we must define how rewards are given. Question: Is it correct to reward the network at every moment in time?\\ 
- Perhaps random activity in a neuronal group at a particular moment might lead to an incorrect reward for the network. Thus, a process is needed to prevent this.\\ 
To solve this issue, rewards are given to the network after it identifies the response. That is, at each moment, all spikes occurring in the network are stored. Once the network's decision is determined, a reward corresponding to this decision is given. But a new question arises: How can we incorporate the network’s confidence in a decision into the reward? Clearly, if the network provides a wrong answer with high confidence, it should be penalized much more than if it gives a wrong answer with low confidence.\\ 
To address this, two approaches are adopted. The first approach, used in the network design, involves calculating the difference in activity between the neuronal groups in the second layer to select the response. To evaluate the network's confidence, the activity difference is measured and used to assess whether the network is confident or uncertain about its response, based on a predefined threshold. However, this introduces a new parameter for learning, complicating the process. To mitigate this, the second approach is executed alongside the first. If the confidence threshold is set reasonably, there is no need for manual adjustment to accurately represent the network's confidence. Instead, for each test input, the input is provided to the network multiple times to obtain several responses. Based on the fraction of correct responses among these, the network's confidence can be assessed. If the response is consistent, the network is considered highly confident; if not, random neuronal activity may cause variation or insufficient activity may result in no definitive response. Once the confidence percentage is calculated, the network is trained accordingly with a reward proportional to this confidence.

\subsubsection{Reward Calculation Formula}
The only remaining part is the reward calculation method. Up to this point, we have kept two pieces of data: one is the spike history, and the other is the network's confidence percentage. One of the most common methods for rewarding or punishing spiking networks is to give a fixed amount of reward. However, as mentioned in the previous section, this approach is not beneficial for us. Therefore, we introduce two novel methods for providing dynamic rewards
\begin{itemize}
	\item [1] Calculating the reward using the number of correct responses in the last ten responses
	\item[2] Calculating the reward using the difference between the number of correct and incorrect responses in the last ten responses
\end{itemize}

\subsubsection{Reward Calculation Using the Number of Correct Responses in the Last Ten Responses}
First, let's remind ourselves that for decision-making in the model, we showed the test inputs to the model multiple times. From this point on, let's assume this number is \textbf{ten times}. Also, let $H_C$ be the number of times the model provided the correct response in the last ten tests, and $H_{IC}$ be the number of times the model provided an incorrect response in the last ten tests. Now, to calculate the reward, we have:
\begin{equation}
	Reward=\frac{H_{IC}}{10},~Punishment=\frac{H_{C}}{10}
\end{equation}

Thus, if the model is very confident in its answer and the connection doesn't contribute to the answer, that connection is heavily punished. This is because the current connections that are helpful in finding the answer have already been sufficient for solving the problem, and there is no need to retain undesirable connections. However, if the model is unsure about its answer, it cannot be confidently said that the connection was either helpful or undesirable. Therefore, in this situation, only a small reward or punishment is applied.

\subsubsection{Reward Calculation Using the Difference Between the Number of Correct and Incorrect Responses in the Last Ten Responses}
The main issue not considered in the previous method is the case where the network is in the early stages of training, and the activity of the neuronal groups is not sufficiently different to confidently determine whether the response to a training input should be positive or negative. To solve this problem, we have developed this method. In this case, we first calculate the reward and punishment using the previous method. However, the network might have, for example, estimated the training input as positive five times and negative five times. For such cases where the confidence is low, the reward and punishment are multiplied by the confidence percentage. Also, if the network’s response to a specific input is not clear, we apply some standard deviation to the input flow of neurons in the network, increasing the randomness of the network’s activity and enabling the network to search for and discover optimal connections.
In addition to defining the variables $H_C$ and $H_{IC}$, which were referenced in equation 5.1, we now introduce a new coefficient, $K$, which represents the network's confidence in the answer obtained from the ten executions. The formula for calculating this coefficient is defined as follows:
\begin{equation*}
	K = \frac{|H_C - H_{IC}|}{10}
\end{equation*}
Thus, the formula for the improved training method becomes:
\begin{equation}
	Reward=K\times\frac{H_{IC}}{10}, Punishment=K\times\frac{H_{C}}{10}
\end{equation}

\section{Input Encoders}
Before introducing the structure of the network used, it is necessary to explain a few points and assumptions in this report.

\subsection{Dataset Used}
For this project, we aimed to use sports video datasets from various sports, asking the model to classify the sports. To achieve this, one of the well-known datasets in this area is the Kinetics-400
\cite{Kineticks400}
dataset, which contains approximately 650,000 videos in 400 sports and activity categories. Each category includes a large number of videos related to the activity, sourced from YouTube. Each video clip in this dataset is about 10 seconds long.
\begin{figure}[h!]
	\begin{center}
		\includegraphics[height=6cm]{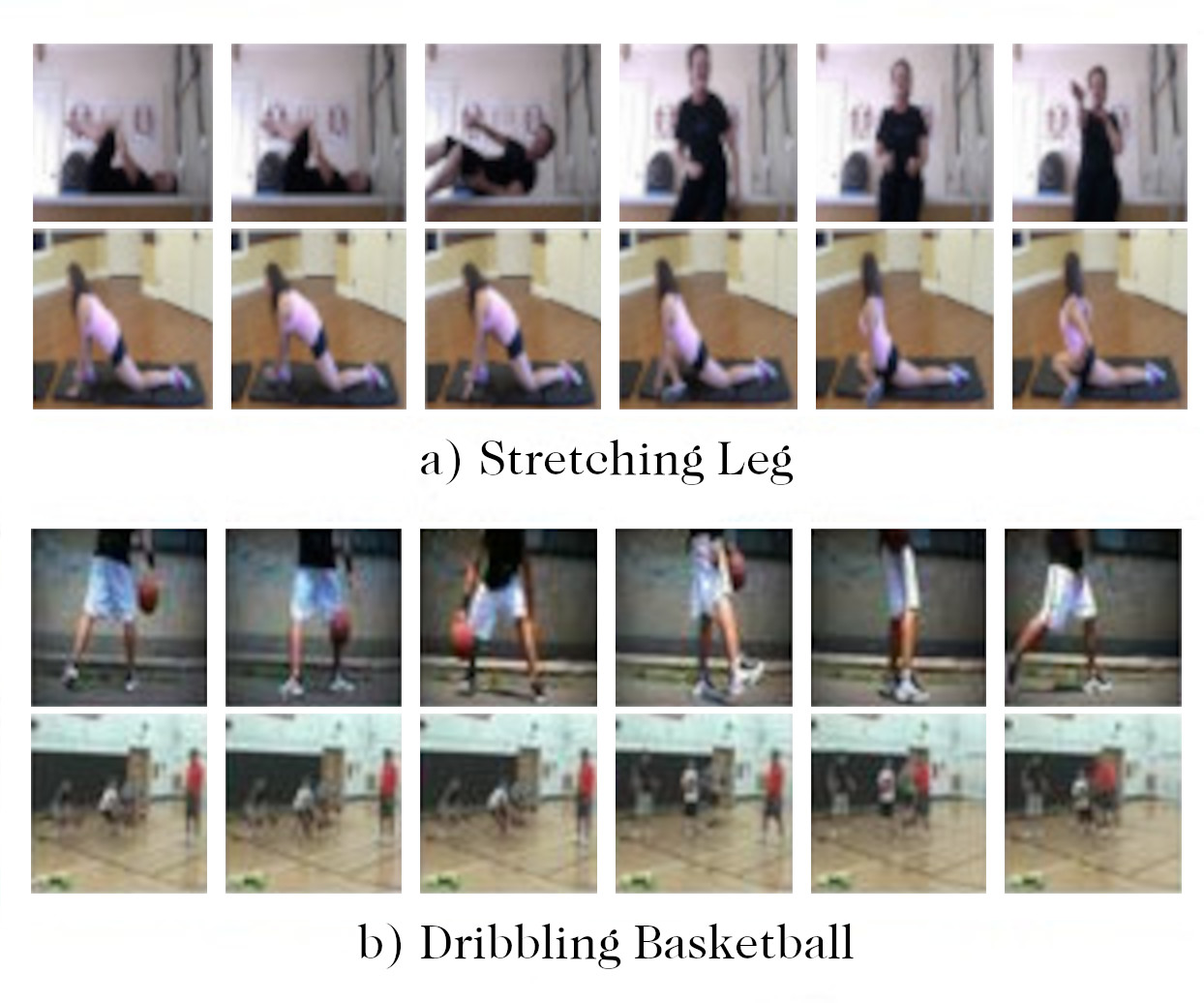}
	\end{center}
	\caption{
		An example of sports videos from the 
		Kinetics-400 dataset.
	}
\end{figure}

\subsection{Encoding Used}
Given the nature of the dataset we are working with and the structure of our network, we need to perform two transformations here:
\begin{itemize}
	\item [1] Convert videos into a series of images
	\item [2] Convert images into a spike train
\end{itemize}
Let's start with the first transformation.

\subsubsection{Converting Video to Images}
Imagine a video of an athlete playing tennis in a stadium. What data can we find in this video?
\begin{itemize}
	\item [1] The first type of data is visual. For example, there is an athlete, there is a tennis court, and so on.
	\item [2] The second type of data, which is highly useful when analyzing videos, is temporal data. This means that the athlete was serving in the previous frame, just completed the serve, and is moving forward preparing for defense in the next frame.
\end{itemize}

\begin{figure}
	\begin{center}
		\includegraphics[height=6cm]{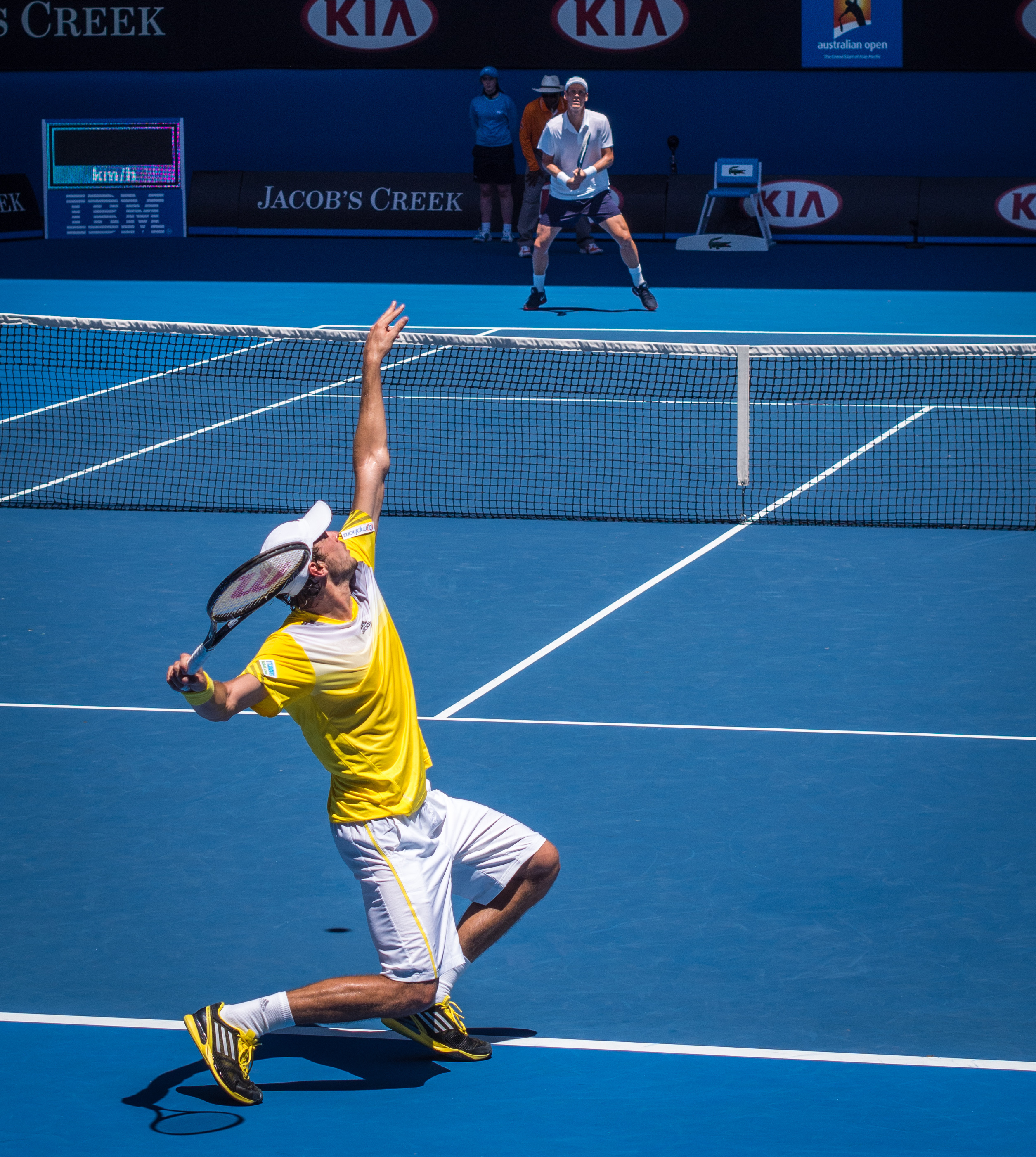}
	\end{center}
	\caption{
		A frame from a tennis match video, containing both visual and temporal data.
	}
\end{figure}

It is clear that analyzing a video requires considering both types of data in it. The first type of data is traditionally used in image processing networks, like the network mentioned in 
\cite{second}, 
and there is a lot of prior knowledge on how to convert it. The main challenge is analyzing the second type of data, since it involves the dimension of time, and classical networks that do not consider time are not capable of processing it. However, this limitation does not exist for us in spiking networks! This is because each spike enters the network after the previous one at a specific time. Therefore, to convert video to images, we simply need to provide each video frame in sequence, separately, with its corresponding timestamp to the network.

Now that we understand how to convert video to images, we will move on to converting the images into a spike train.

\subsubsection{Converting Image to Spike Train}
For this task, inspired by
\cite{second},
we propose several methods:
\begin{itemize}
	\item [5.2.3] Time to first spike
	\item [5.2.4] Poisson distribution
	\item [5.2.5] Position encoding with a fixed dictionary
	\item [5.2.5] Separate position and word encoding
	\item [5.2.6] Random corresponding encoding set
	\item [5.2.7] Separate position and word encoding with Gaussian distribution
\end{itemize}

For simplicity in explaining this section, we refer to the distinct value of each pixel as a word, and the linear chain of pixels in an image as a sentence. For convenience, let's assume the images are black and white, so their color is one-dimensional.

\subsubsection{Time To First Spike}
This is an encoding that has been used in many projects however since it takes many assumptions about the input which may or may not be correct, we will not use it in our project. But, for the sake of completeness we will explain and show a demo.

This method works by mapping an image into grayscale, and taking their pixel values. It then creates the encoding based on the value in the range. The pixels can be from $0$ to $255$. If we define the time-window for the encoding as $window$, a pixel with value $pixel$ will spike at time $t$ where:
\begin{equation*}
	t = window \times \frac{pixel}{(255-0)}
\end{equation*}
An example of this encoding is given below.

\begin{figure}[h!]
	\begin{center}
		\subfigure[Bridge]{
			\includegraphics[width=0.17\textwidth]{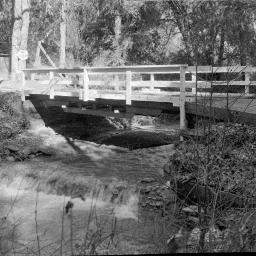}
		}
		\subfigure[Bird]{
			\includegraphics[width=0.17\textwidth]{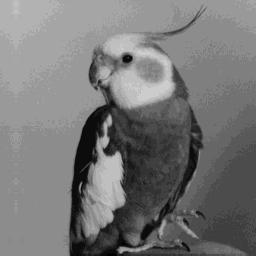}
		}\\
		\subfigure[Encoded Bridge]{
			\includegraphics[width=0.17\textwidth]{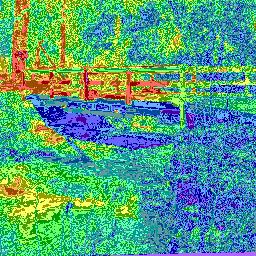}
		}
		\subfigure[Encoded Bird]{
			\includegraphics[width=0.17\textwidth]{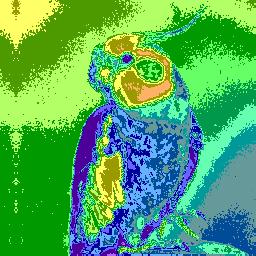}
		}\\
		\subfigure[Intensity]{
			\includegraphics[width=0.3\textwidth]{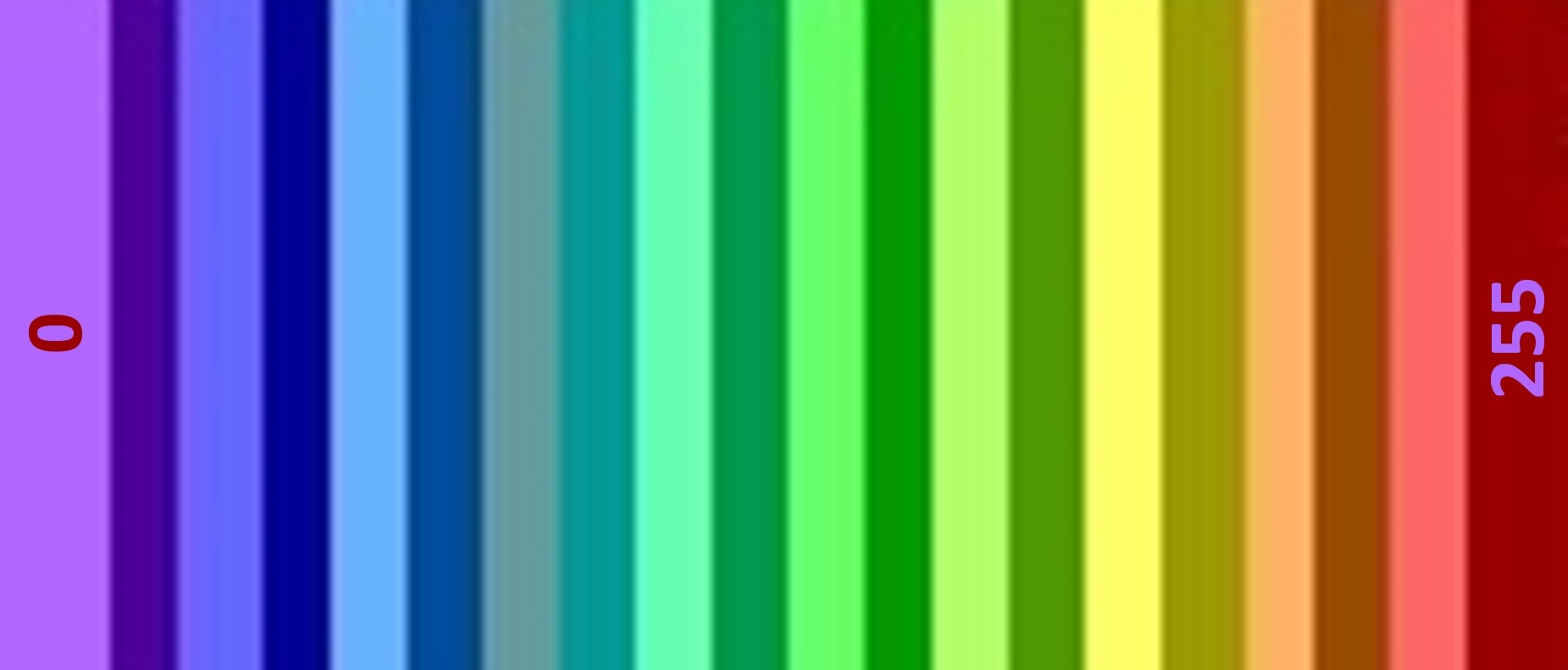}
		}
	\end{center}
	\caption{
		A sample representation of two images from \cite{WaterlooDataset}, and their respective encodings' heat maps using Time-To-First-Spike encoding. Pixles with values close to 255 fire closer to 0ms.
	}
\end{figure}

\begin{figure}[h!]
	\begin{center}
		\subfigure[Bridge]{
			\includegraphics[width=0.22\textwidth]{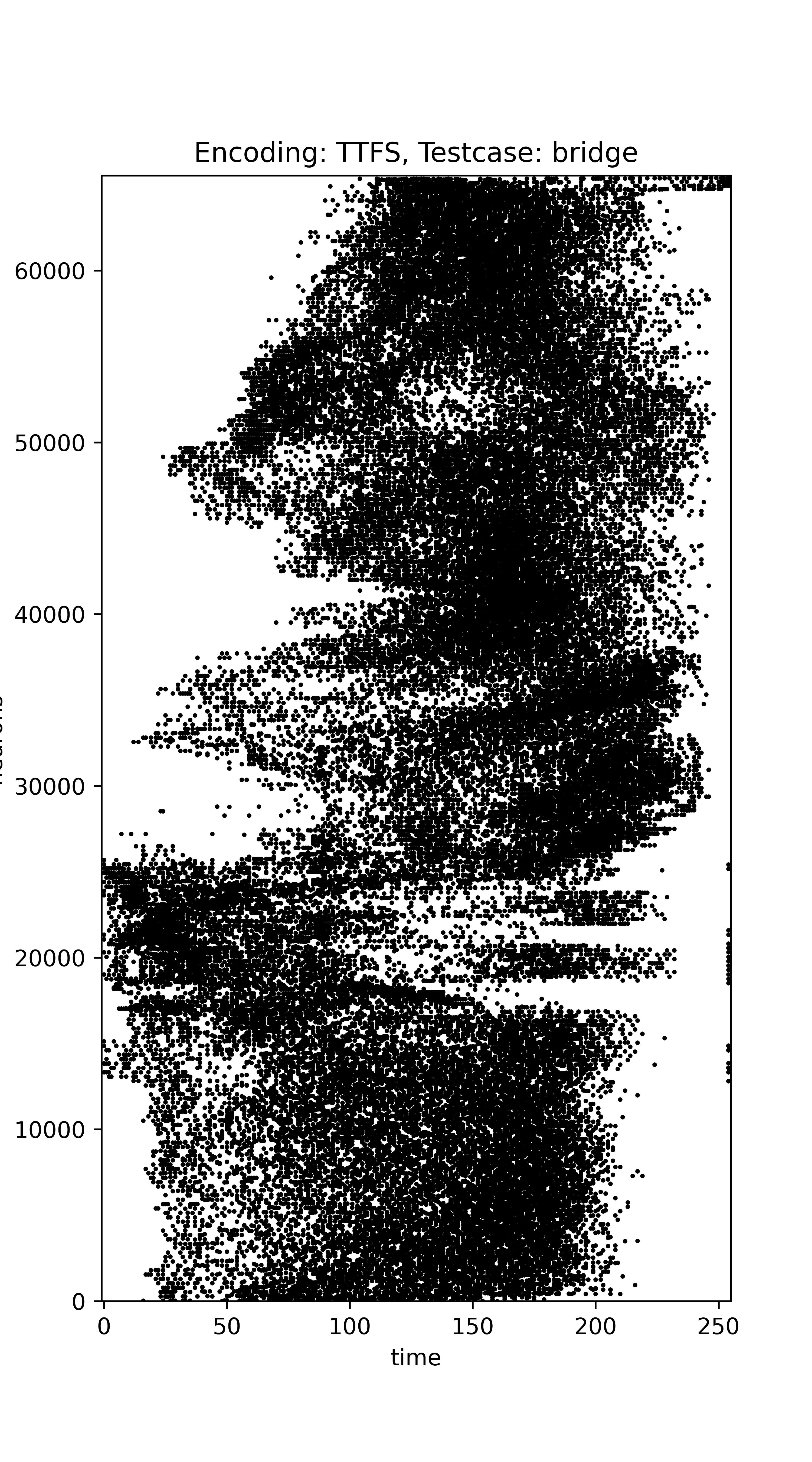}
		}
		\subfigure[Bird]{
			\includegraphics[width=0.22\textwidth]{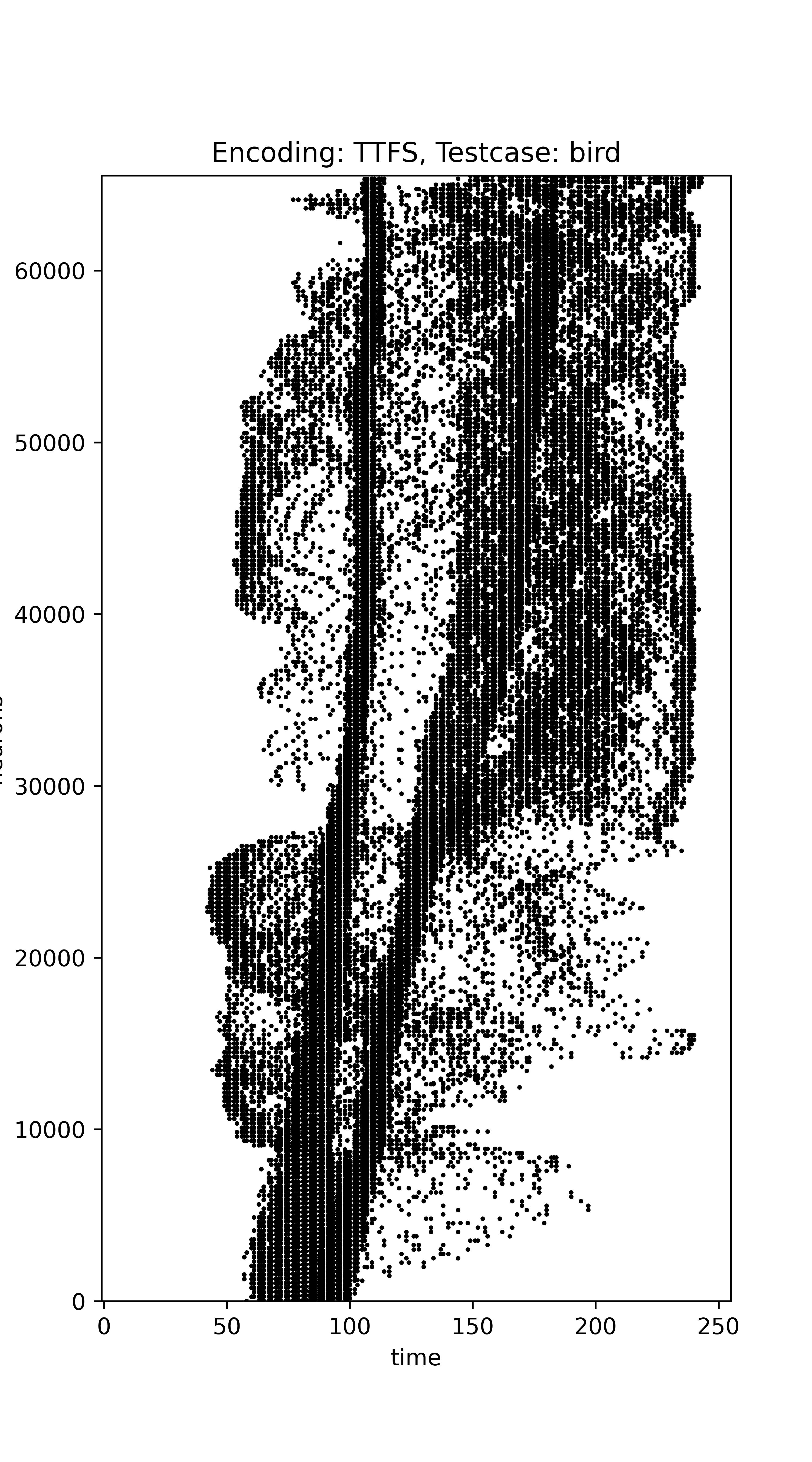}
		}\\
	\end{center}
	\caption{
		The corresponding spike train encoding to the previous heat maps.
	}
	\label{fig3}
\end{figure}

\subsubsection{Poisson Distribution}
In this encoding model, we first consider a fixed dictionary of the most frequently occurring values. Then, for each word in the dictionary that appears in the input, we provide random activity following a Poisson distribution to the neuron corresponding to that word.

\begin{figure}
	\begin{center}
		\includegraphics[height=8cm]{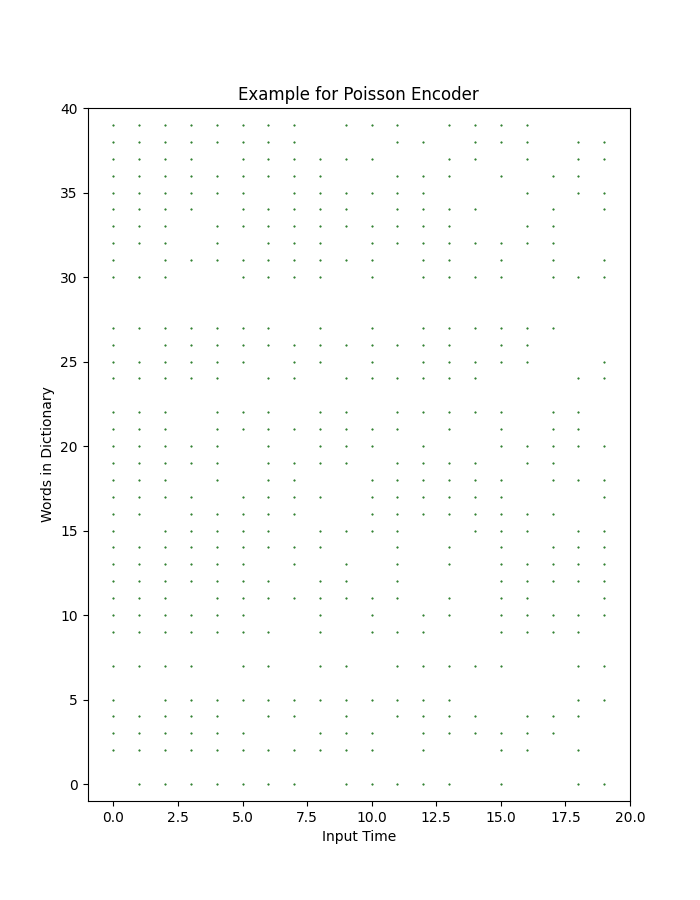}
	\end{center}
	\caption{
		Example of encoding input data using the Poisson distribution
	}
\end{figure}

\subsubsection{Position Encoding Using a Fixed Dictionary}
In this encoding model, we behave similarly to the previous case. The difference is that instead of providing random activity to a word from the dictionary that appears in the input, we consider an interval of arbitrary length corresponding to the length of the input sentence, and place the activity at the relative position of the word within the input.

\begin{figure}
	\begin{center}
		\includegraphics[height=8cm]{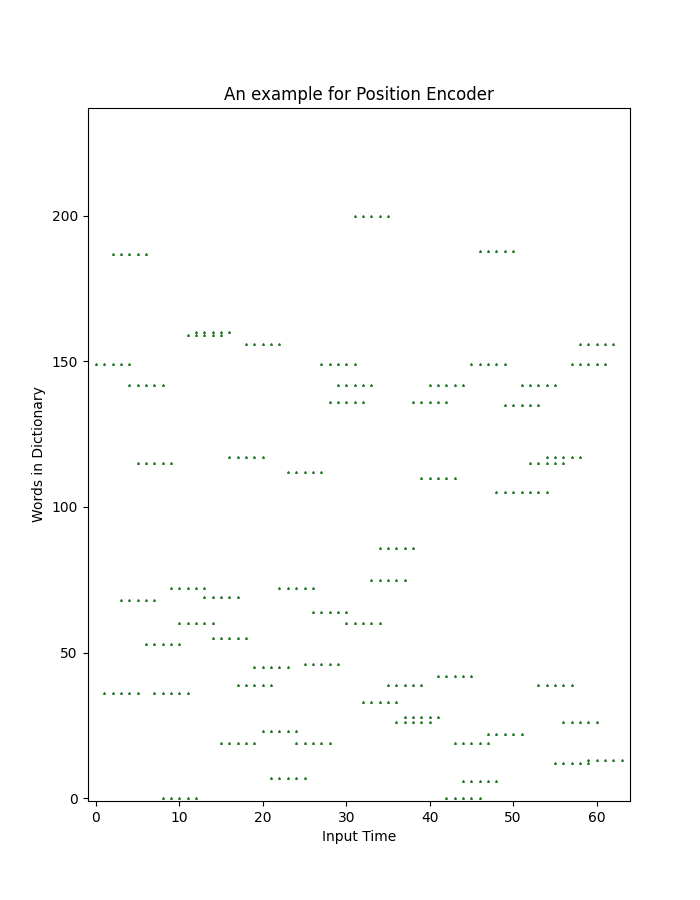}
	\end{center}
	\caption{
		Example of encoding input data using position encoding with a fixed dictionary
	}
\end{figure}

\subsubsection{Separate Position and Word Encoding}
Unlike the previous methods, here we separate the encoding of the word's position and the word's concept. For each word in the dictionary, we assign a neuron. Additionally, for an interval of arbitrary length, which represents the length of the sentence, we assign a neuron corresponding to the position. Then, for each word in the dictionary that appears in the sentence, both the corresponding word neuron and the neuron corresponding to its relative position spike simultaneously.

\begin{figure}
	\begin{center}
		\includegraphics[height=7cm]{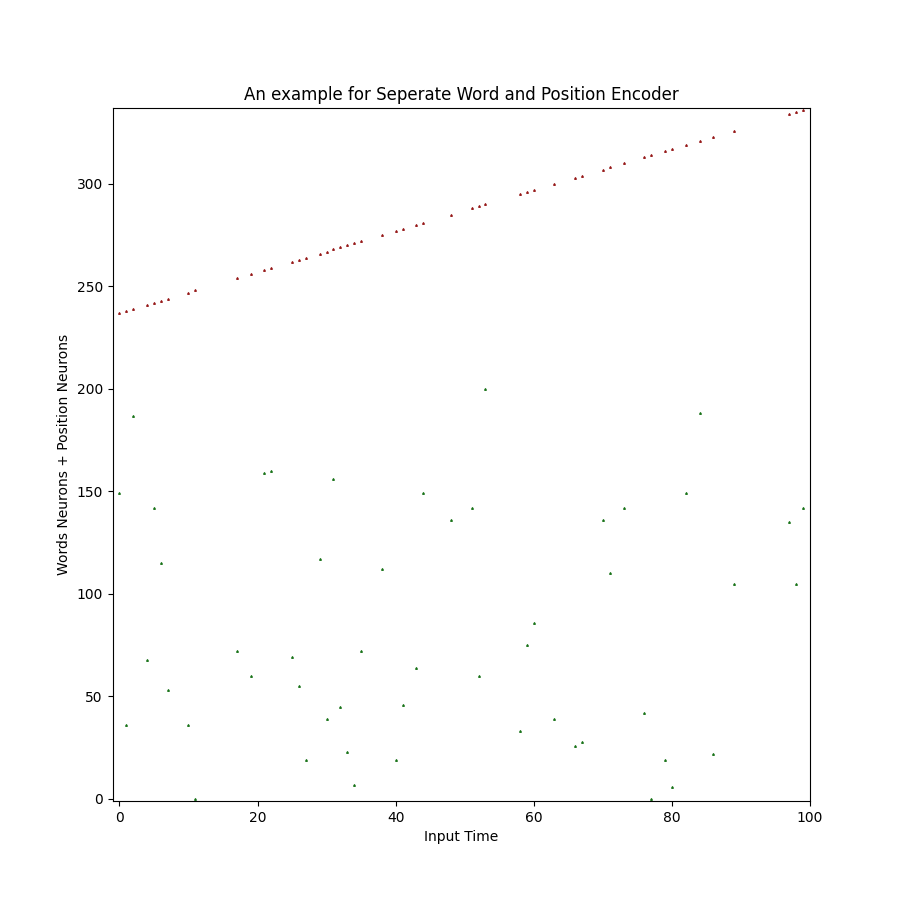}
	\end{center}
	\caption{
		Example of encoding input data using separate position and word encoding
	}
\end{figure}

\subsubsection{Random Corresponding Encoding Set}
This method is a combination of the first and third methods. In this approach, we consider several neurons corresponding to words and several corresponding to the positions of the words. Then, using the Poisson distribution, for each word in the dictionary and each position, we create an invariant encoding in which $10\%$ of the neurons are randomly selected to spike. Now, to convert each input at any given moment, we consider the encodings created for the word in the input and the position of the word.

\begin{figure}
	\begin{center}
		\includegraphics[height=8cm]{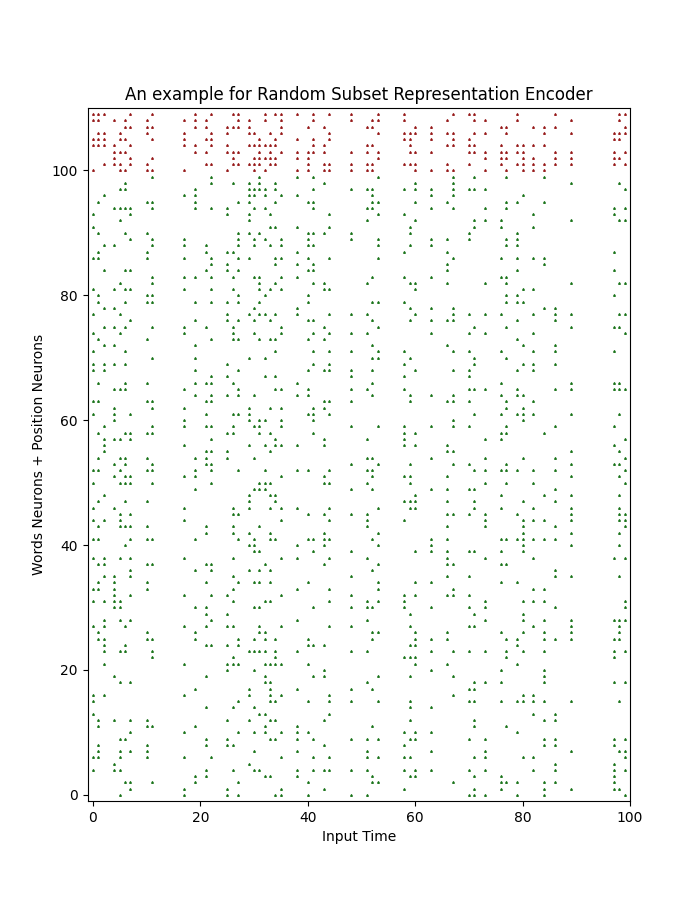}
	\end{center}
	\caption{
		Example of encoding input data using the random corresponding encoding method
	}
\end{figure}

\subsubsection{Separate Position and Word Encoding with Gaussian Distribution}
In the final method, for encoding words, we use the previous method, but for encoding the position, we create a random distribution with a mean corresponding to the relative position of the word in the sentence; and from that, we randomly select several samples to be the neurons that should spike. This method is similar to the third method, but instead of selecting the exact position for the spike, a position close to it is selected.

\begin{figure}
	\begin{center}
		\includegraphics[height=7cm]{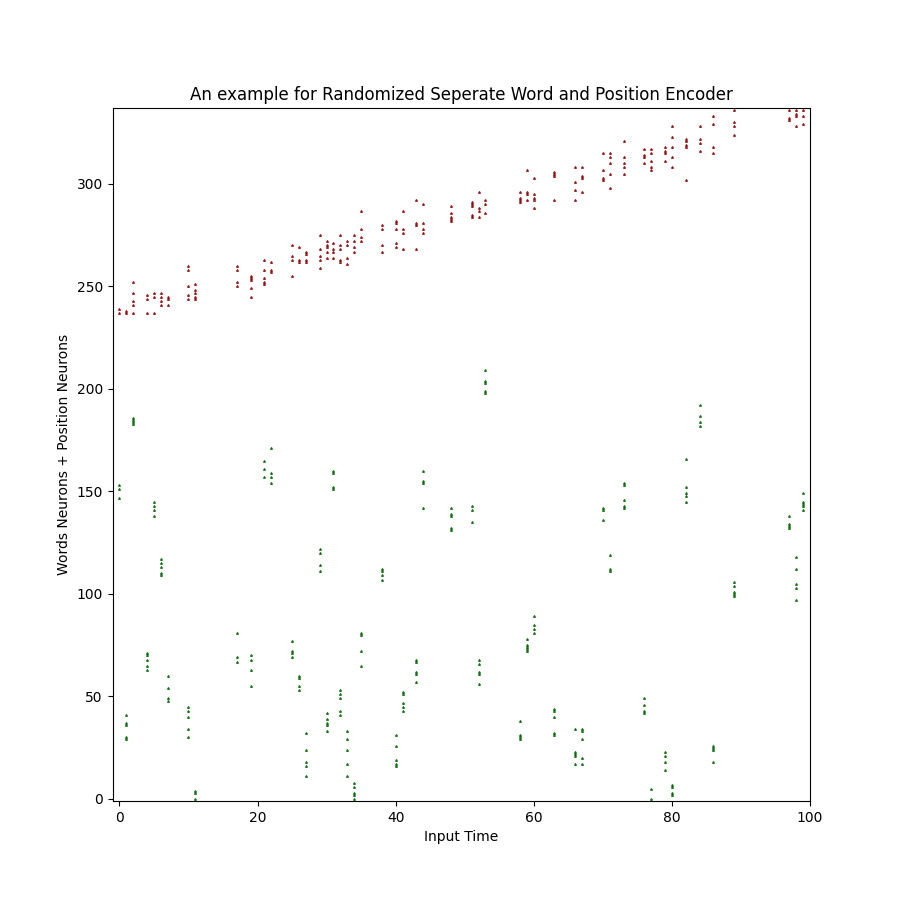}
	\end{center}
	\caption{
		Example of encoding input data using separate position and word encoding with Gaussian distribution
	}
\end{figure}
Finally, after designing these five novel methods for generating spike train encodings, we used method 4 for convenience in our experiments.

\subsection{Corresponding Dataset}
So far, the inputs of our dataset have been converted into spike train sequences. However, we know that processing images is a time-consuming and costly task; to conduct numerous experiments, we need to propose an innovative approach to reduce their time and cost. For this purpose, we are inspired by
\cite{second}.
\\
Note that initially, transformer models were built for text processing, and when applied to image processing, innovative methods were proposed for interpreting an image like text. One of these methods was dividing an image into smaller patches and treating each patch as a word.

The transformer model, which previously worked only with words, now had the capability to work with images. Thus, following a similar approach, in our experiments, instead of using images that have hundreds of pixels and are very costly, we use several sentences. With the encoding methods explained in the previous section, it is clear that the method for converting an image to a spike train can also be applied to words in sentences. For example, the encodings that were shown earlier were all encodings corresponding to the following phrase with 256 words. We can assume that each word represents a black-and-white pixel, and repeated words represent repeated values for colors, in which case the following phrase would correspond to an image of size $16 \times 16$.\\

\textcolor{gray}{
	The film I watched last night was a mix of good and bad elements, making it a fascinating watch. The plot centered around a young man struggling with life’s challenges. While the story had potential, it was poorly executed, with some parts even feeling boring. The acting, however, was excellent in some scenes, with wonderful performances that brought the characters to life. Still, there were moments of terrible, awful acting that made me cringe. It’s funny how a film can have both the best and worst qualities at the same time. I always like a good story, but this one had nothing particularly new to offer. The beautiful cinematography was perfect in capturing the essence of the scenes, yet the overall direction felt stupid and misguided at times. The family dynamics were well-depicted, adding depth to the plot. However, some of the acting was so poor that it detracted from the film’s potential impact. There were parts that I really liked, and I could see the effort to make something memorable, but it also had some horrible moments. It’s a waste to see such good talent put to such bad use. Even with its flaws, there were at least a few scenes that were excellent. I think it could have been better with a more focused storyline and better direction. The film had its wonderful moments but was overall a disappointing experience. Life is too short to spend on movies that don’t deliver, but this one wasn’t the worst I’ve seen. It’s not the best, period.
}

So, for the experiments, we design a smaller corresponding dataset to simplify our computations. In this dataset, there are 100 sentences with a maximum length of order 20, and for each sentence, there is a response indicating whether the sentence has a positive sentiment or not. Now, what the model has to do for classification is, for each sentence it encounters in the input, to create the expected spike train encoding for that sentence and then attempt to classify it as either positive or negative.
Some examples from the dataset can be found in Table \ref{tab:dataset}.
\begin{table}[h!]
    \caption{Examples from the sentiment dataset}
	\begin{tabular}{ccl}
		\hline 
		Review & Sentiment\tabularnewline
		\hline 
		\hline 
		The humor was forced, not really funny. Disappointed. & neg\tabularnewline
		\hline 
		Stunning animation brought the characters to life! & pos\tabularnewline
		\hline 
		Spectacular visuals but the plot fell flat. Unimpressed. & neg\tabularnewline
		\hline
		Engaging from start to finish, highly recommend it. &pos\tabularnewline
		\hline
		Slow pacing made it hard to stay interested. Boring. & neg\tabularnewline
		\hline
		A brilliant twist ending that I didn't see coming. & pos\tabularnewline
		\hline
		The plot holes were too glaring to ignore. Poor. & neg\tabularnewline
		\hline
		Excellent direction and superb acting. A must-watch! & pos\tabularnewline
		\hline
		Terrible dialogue, felt very unnatural. Bad writing. & neg\tabularnewline
		\hline
		Intriguing plot with unexpected twists. Great film! & pos\tabularnewline
		\hline
	\end{tabular}
    \label{tab:dataset}
\end{table}

\section{Experiments}
Now, using the dataset and encoding introduced in Chapter 3, the model from Chapter 4, and the learning procedure from Chapter 5, we aim to evaluate the effectiveness of this network in learning the features of input spike train sequences.

\subsection{Single Input}
We start with a simple case. In this case, initially, we trained the network with a large number of positive input samples. The weights of the connections for the neuron group corresponding to the positive response were approximately five times greater than the weights for the neuron group corresponding to the negative response. Now, we want to present a negative input and check over time whether the network can learn or not. We performed this experiment multiple times, and below we show the responses where the network had the highest and lowest confidence in its outputs.

\begin{figure}[h!]
	\begin{center}
		\subfigure[Lowest Confidence]{
			\includegraphics[width=0.4\textwidth]{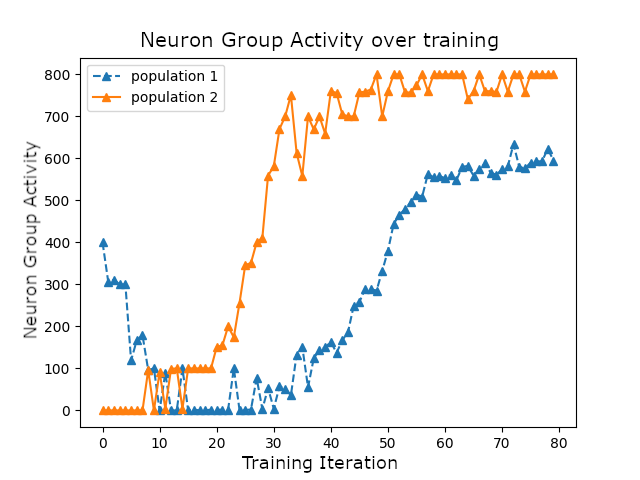}
		}\\
		\subfigure[Highest Confidence]{
			\includegraphics[width=0.4\textwidth]{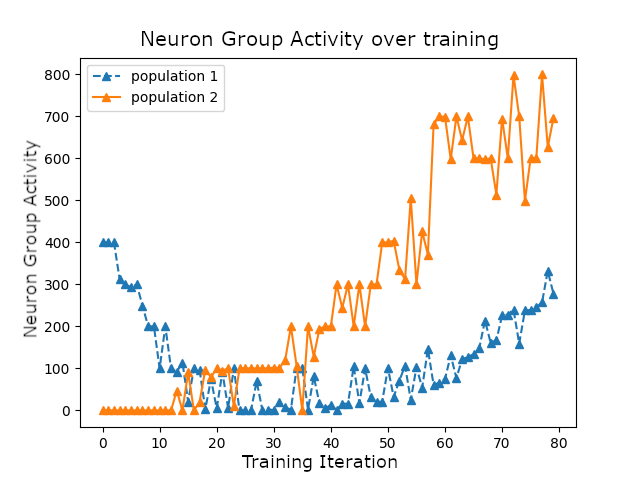}
		}
	\end{center}
	\caption{
		Network's learning from previously trained positive inputs when presented with a new negative input
	}
\end{figure}

\subsection{Four Inputs}
Now that we are confident in the network's functionality, we examine a case where the network sees a larger but limited number of inputs. In this approach, we proceed similarly to the previous case, except that instead of a single new input, we present four new inputs to the network. To demonstrate the network's performance, we use the activity of the neuron groups; this time, we consider the group corresponding to our input's response as the first group and the competing group as the second group. We can observe that consistently, the network is able to distinguish between the inputs and learn them, classifying them correctly in each run.
\begin{figure}[h!]
	\begin{center}
		\subfigure[First Experiment]{
			\includegraphics[width=0.4\textwidth]{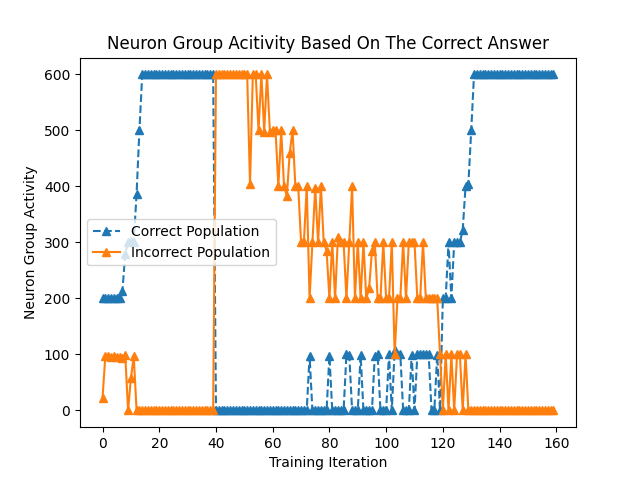}
		}\\
		\subfigure[Tenth Experiment]{
			\includegraphics[width=0.4\textwidth]{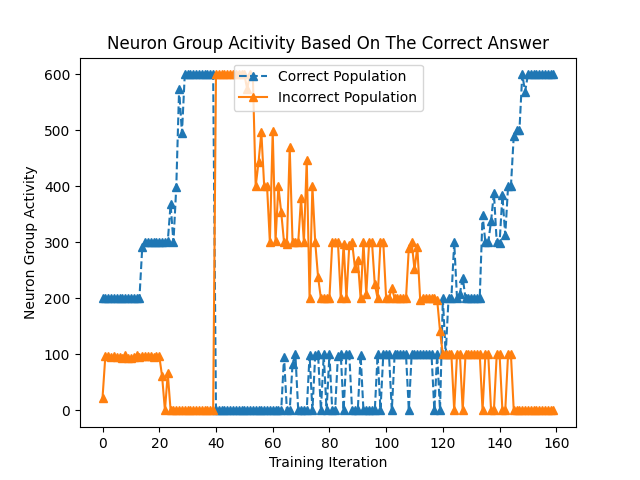}
		}
	\end{center}
	\caption{
		Network's learning from previously trained positive inputs when presented with four new inputs
	}
\end{figure}

\subsection{Multiple Inputs}
In the final case, we increase the number of inputs to ten to see how the network behaves when faced with many inputs that share similar elements. The experiment in this section is exactly like the previous one, except that the number of inputs has increased.
\begin{figure}[h!]
	\begin{center}
		\subfigure[First Experiment]{
			\includegraphics[width=0.4\textwidth]{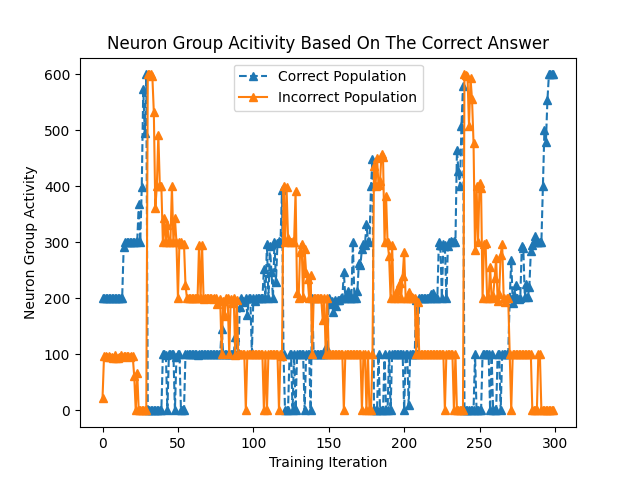}
		}\\
		\subfigure[Third Experiment]{
			\includegraphics[width=0.4\textwidth]{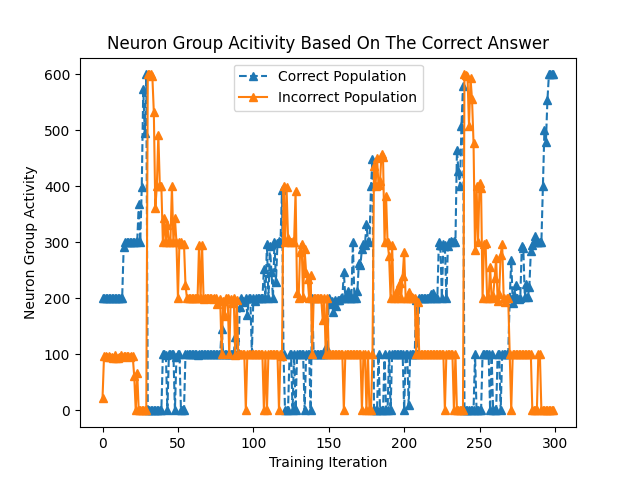}
		}
	\end{center}
	\caption{
		Network's learning from previously trained positive inputs when presented with ten new inputs
	}
\end{figure}

To improve the results, we first note that the human brain cannot learn a large number of input data at a quick glance. For example, a newborn child cannot speak on their first attempt; they must be repeatedly encouraged and taught by their parents to learn how to speak. Therefore, with this logic, we similarly hypothesize that to improve results, we should increase the number of times the data is seen.
\begin{figure}[h!]
	\begin{center}
		\includegraphics[height=14cm]{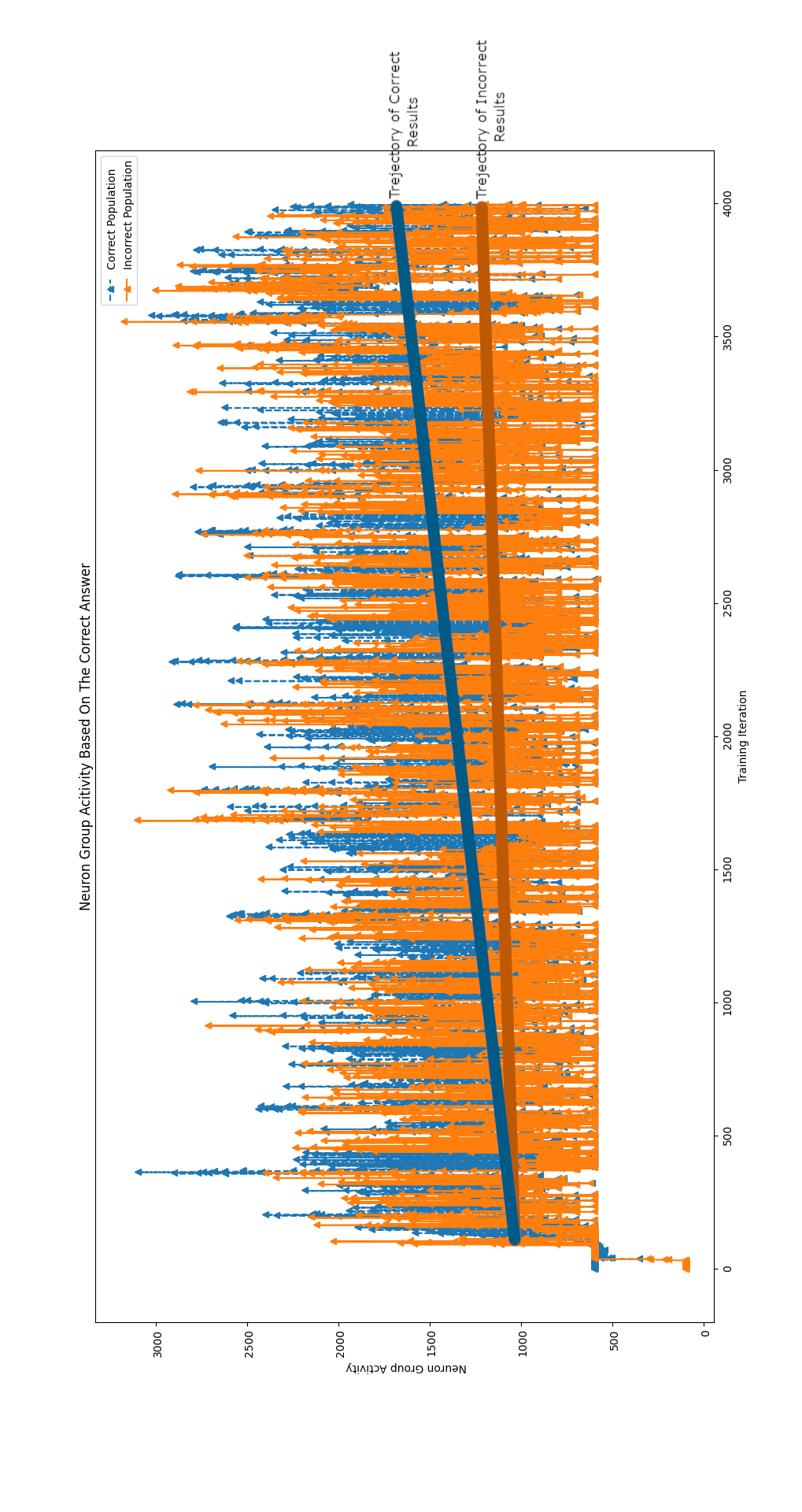}
	\end{center}
	\caption{
		Network's learning from previously trained positive inputs when presented with ten new inputs and multiple repetitions
	}
	\label{fig4.13}
\end{figure}

\section{Conclusion}
In this work, we studied hybrid models, and introduced a promising alternative. In particular, we introduced a new architecture to analyze time-series data. Inspired by cortical column structures in the brain, this novel architecture bridges the gap between temporal and spatial data processing. We also introduced several encoding methods one might use for this model and discussed visual and textual input data. Using the said encoders, we designed a training procedure for the network. And showed its results in a few experiments. The proposed model demonstrates its potential for more efficient classifications while maintaining the same level of accuracy, particularly in smaller-scale applications. Finally, we published our complete code implementations to encourage future research in this field. This work underscores the viability of biologically inspired computing paradigms and paves the way for advancements in computational neuroscience and machine learning.

\section{Future Work}
As the experiments show, our spiking neural network performs well with small inputs, but the main issue arises when the amount of data increases significantly. In this case, training the network becomes a time-consuming process, and the learning performance of the network decreases compared to classical networks. As you can see in Fig. 14, in which the orange points represent incorrect responses and the blue points represent correct responses from the network, the blue points ultimately show an increasing trend, eventually surpassing the orange points. This indicates that the network successfully learned the output. However, the problem here is that this process is very slow, and it took hours of training. Such time expenditure is not practical for training neural networks in daily operations.

The reason for the long learning time is that in classical networks, whether the model is learning from one sample or multiple samples, it spends a fixed amount of time on each sample. However, in these networks, each input sample must be shown multiple times to the network so that the time-dependent spike-time dependent plasticity algorithm can strengthen the synaptic connections between the pre-synaptic neurons and post-synaptic neurons at all favorable spike times. The problem arises when this algorithm weakens the synaptic connections. This happens when multiple input samples are shown to the network, and while learning the output, the network may forget part of the previous sample. As a result, as the number of input samples increases, the number of times each input sample must be seen also increases significantly for the network to learn the pattern properly.

These results indicate the great potential of networks inspired by the structure of the cortical column of the brain, especially the network presented in this report in current small-scale infrastructures. Not only do these networks benefit from fewer computations and, as a result, lower power consumption, but they also have high capabilities in handling time. However, for generalizing to larger structures, more complex and powerful hardware must be used for training, which results in the networks performing worse than classical networks. Therefore, for now, it is reasonable to use them only in small-scale structures. The main reason for this is the difficulty of training many patterns on these networks. When the number of input patterns becomes very large, the network requires much more time to learn them.
In future research by the scientific community, it would be beneficial to explore how we can improve the learning speed of these networks. The results obtained show spiking networks as a promising alternative. Even in small scales, we have achieved significant success. However, it seems that the key to controlling these networks at large scales will be improving their learning performance.

\bibliographystyle{ACM-Reference-Format}
\bibliography{references}

\end{document}